\documentclass[11pt]{article}
\usepackage{acl2016}
\usepackage{times}
\usepackage{latexsym}
\usepackage{multirow}
\usepackage{amsmath}
\usepackage{amsfonts}
\usepackage{url}
\usepackage{graphicx}
\usepackage{float}
\usepackage{afterpage}
\usepackage{hyperref}

\aclfinalcopy 

\title{Generating Factoid Questions With Recurrent Neural Networks: \\ The 30M Factoid Question-Answer Corpus}

\author{Iulian Vlad Serban${^{*}}{^{\circ}}$
       \And Alberto Garc\'ia-Dur\'an${^{*}}{^{\diamond}}$
       \And Caglar Gulcehre${^{\circ}}$
       \And Sungjin Ahn${^{\circ}}$
       \AND Sarath Chandar${^{\circ}}$
       \And Aaron Courville${^{\circ}}$
       \And Yoshua Bengio${^{\dagger}}{^\circ}$}

\date{}

\newcommand\blfootnote[1]{%
  \begingroup
  \renewcommand\thefootnote{}\footnote{#1}%
  \addtocounter{footnote}{-1}%
  \endgroup
}

\begin{document}
\maketitle

\begin{NoHyper}
\blfootnote{* First authors.}
\blfootnote{${^{\circ}}$ University of Montreal, Canada \\
Email: \{iulian.vlad.serban,caglar.gulcehre,sungjin.ahn,\\
sarath.chandar.anbil.parthipan,aaron.courville,yoshua.bengio\}@umontreal.ca
}
\blfootnote{${^{\diamond}}$ Universit\'e de Technologie de Compi\`egne - CNRS, France \\
Email: alberto.garcia-duran@utc.fr}
\blfootnote{${^{\dagger}}$ CIFAR Senior Fellow}
\end{NoHyper}

\begin{abstract}
Over the past decade, large-scale supervised learning corpora have enabled machine learning researchers
to make substantial advances.
However, to this date, there are no large-scale question-answer corpora available.
In this paper we present the 30M Factoid Question-Answer Corpus,
an enormous question answer pair corpus produced by applying
a novel neural network architecture on the knowledge base Freebase 
to transduce facts into natural language questions.
The produced question answer pairs are evaluated both by human evaluators and 
using automatic evaluation metrics, including well-established machine translation and sentence similarity metrics.
Across all evaluation criteria the question-generation model outperforms the competing template-based baseline.
Furthermore, when presented to human evaluators, the generated questions appear comparable in quality to real human-generated questions. 
\end{abstract}
\section{Introduction}
A major obstacle for training question-answering (QA) systems has been due to the lack of labeled data. 
The question answering field has focused on building QA systems based on traditional information retrieval procedures \cite{lopez2011question,dumais2002web,voorhees2000overview}.
More recently, researchers have started to utilize large-scale knowledge bases (KBs)
\cite{lopez2011question}, such as Freebase~\cite{bollacker2008freebase}, WikiData~\cite{vrandevcic2014wikidata} and Cyc~\cite{lenat1989building}.\footnote{Freebase is now a part of WikiData.}
Bootstrapping QA systems with such structured knowledge is clearly beneficial,
but it is unlikely alone to overcome the lack of labeled data.
To take into account the rich and complex nature of human language,
such as paraphrases and ambiguity,
it would appear that labeled question and answer pairs are necessary.
The need for such labeled pairs is even more critical for training neural network-based QA systems,
where researchers until now have relied mainly on hand-crafted rules
and heuristics to synthesize artificial QA corpora~\cite{bordes2014QA,bordes2015large}.

Motivated by these recent developments, in this paper we focus on generating 
questions based on the Freebase KB. We frame question generation as a transduction 
problem starting from a Freebase fact, represented by a triple consisting of a
subject, a relationship and an object, which is transduced into a question about 
the subject, where the object is the correct answer \cite{bordes2015large}.
We propose several models, largely inspired by recent neural machine translation models
\cite{Cho-et-al-EMNLP2014,sutskever2014sequence,bahdanau2014neural},
and we use an approach similar to Luong et al.\@ \shortcite{luong2015addressing} for dealing with the problem of rare-words.
We evaluate the produced questions in a human-based experiment as well as with respect to automatic evaluation metrics, 
including the well-established machine translation metrics BLEU and METEOR and a sentence similarity metric.
We find that the question-generation model outperforms the competing template-based baseline,
and, when presented to untrained human evaluators, the produced questions appear to be indistinguishable from real human-generated questions.
This suggests that the produced question-answer pairs are of high quality and therefore that they will be useful for training QA systems.
Finally, we use the best performing model to construct a new factoid question answer corpus
-- The 30M Factoid Question-Answer Corpus --
which is made freely available to the research community.\footnote{\url{www.agarciaduran.org}} 


\section{Related Work}

Question generation has attracted interest in recent years with notable work by Rus et al.\@ \shortcite{rus2010first}, 
followed by the increasing interest from the Natural Language Generation (NLG) community. 
A simple rule-based approach was proposed in different studies as \textit{wh-fronting} or \textit{wh-inversion} \cite{kalady2010natural,ali2010automation}. 
This comes at the disadvantage of not making use of the semantic content of words apart from their syntactic role. 
The problem of determining the \textit{question type} (e.g. that a \textit{Where-question} should be triggered for locations), 
which requires knowledge of the category type of the elements involved in the sentence,
has been addressed in two different ways: by using named entity recognizers \cite{mannem2010question,yao2010question} or semantic role labelers \cite{chen2009aist}.
In Curto et al.\@ \shortcite{curto2012question} questions are split into classes according to their syntactic structure, 
prefix of the question and the category of the answer, and then a pattern is learned to generate questions for that class of questions. 
After the identification of key points, Chen et al.\@ \shortcite{chen2009aist}
apply handcrafted-templates to generate questions framed in the right target expression by following the analysis of Graesser et al.\@ \shortcite{graesser1992quest}, 
who classify questions according to a taxonomy consisting of 18 categories.

The works discussed so far propose ways to map unstructured text to questions. This implies a 
two-step process: first, transform a text into a symbolic representation (e.g.\@ a syntactic representation of the sentence), 
and second, transform the symbolic representation of the text into the question \cite{yao2012semantics}. 
On the other hand, going from a symbolic representation (structured information) to a question, as we will describe in the next section,
only involves the second step. Closer to our approach is the work by Olney et al.\@ \shortcite{olney2012question}. 
They take triples as input, where the edge relation defines the question template and the head of the triple replaces the placeholder token in the selected question template. In the same spirit, Duma et al.\@ \shortcite{duma2013generating} generate short descriptions from triples by using templates defined by the relationship and replacing accordingly the placeholder tokens for the subject and object. 

Our baseline is similar to that of Olney et al. \shortcite{olney2012question}, where a set of relationship-specific templates are defined. These templates include placeholders to replace the string of the subject. The main difference with respect to their work is that our baseline does not explicitly define these templates. Instead, each relationship has as many templates as there are different ways of framing a question with that relationship in the training set. 
This yields more diverse and semantically richer questions by effectively taking advantage of the fact-question pairs, which Olney et al.\@ did not have access to in their experiments.

Unlike the work by Berant and Liang \shortcite{berant2014semantic}, which addresses the problem of deterministically generating a set of candidate logical forms with a canonical realization in natural language for each, our work addresses the inverse problem: given a logical form (fact) it outputs the associated question.

It should also be noted that recent work in question answering have used simpler rule-based and template-based approaches to generate synthetic questions to address the lack of question-answer pairs to train their models \cite{bordes2014QA,bordes2015large}.

\section{Task Definition}
\subsection{Knowledge Bases}

In general, a KB can be viewed as a multi-relational graph, which consists of a set of nodes (entities) and a set of edges (relationships) linking nodes together. 
In Freebase~\cite{bollacker2008freebase} these relationships are directed and always connect exactly two entities.
For example, in Freebase the two entities \textit{fires\_creek} and \textit{nantahala\_national\_forest} are linked together by the relationship \textit{contained\_by}.
Since the triple \{\textit{fires\_creek}, \textit{contained\_by}, \textit{nantahala\_national\_forest}\} represents a complete and self-contained piece of information,
it is also called a \textit{fact} where \textit{fires\_creek} is the subject (head of the edge), \textit{contained\_by} is the relationship and 
\textit{nantahala\_national\_forest} is the object (tail of the edge).

\subsection{Transducing Facts to Questions}

We aim to transduce a fact into a question, such that:
\begin{enumerate}
\item The question is concerned with the subject and relationship of the fact, and 
\item The object of the fact represents a valid answer to the generated question.
\end{enumerate}
We model this in a probabilistic framework as a directed graphical model:
\begin{align}
P(Q|F) = \prod_{n=1}^N P(w_n | w_{<n}, F),
\end{align}
where $F = (subject, relationship, object)$ represents the fact, 
$Q = (w_1, \dots, w_N)$ represents the question as a sequence of tokens 
$w_1, \dots, w_N$, 
and $w_{<n}$ represents all the tokens generated before token $w_n$.
In particular, $w_N$ represents the question mark symbol '?'.

\subsection{Dataset}

We use the SimpleQuestions dataset \cite{bordes2015large} in order to train our models.
This is by far the largest dataset of question-answer pairs created by humans based on a KB.
It contains over 100K question-answer pairs created by users on Amazon Mechanical Turk\footnote{\url{www.mturk.com}}
in English based on the Freebase KB.
In order to create the questions, human participants were shown one whole Freebase fact at a time
and they were asked to phrase a question such that the object of the presented fact becomes the answer of the question.\footnote{It is not necessary for the object to be the only answer, but it is required to be one of the possible answers.}
Consequently, both the subject and the relationship are explicitly given in each question.
But indirectly characteristics of the object may also be given since the humans have an access to it as well.
Often when phrasing a question the annotators tend to be more informative about the target object by giving specific 
information about it in the question produced.
For example, in the question \textit{What city is the American actress X from?} the city name given in the object informs the human participant that it was in America
- information, which was not provided by either the subject or relationship of the fact.
We have also observed that the questions are often ambiguous: that is, one can easily come up with several possible answers 
that may fit the specifications of the question. Table \ref{tab:stats} shows statistics of the dataset.

\begin{table}
\centering
\small
\begin{tabular}{| c | c | c | c |}
\hline
Questions & Entities & Relationships & Words \\\hline
108,442 & 131,684 & 1,837 & $\sim$77k \\
\hline
\end{tabular}
\caption{Statistics of SimpleQuestions}\label{tab:stats}
\end{table}

\section{Model}
We propose to attack the problem with the models inspired by the recent success of neural machine translation models \cite{sutskever2014sequence,bahdanau2014neural}.
Intuitively, one can think of the transduction task as a ``lossy translation'' from structured knowledge (facts) to human language (questions in natural language), where certain aspects of the structured knowledge is intentionally left out (e.g.\@ the name of the object).
These models typically consist of two components: an encoder, which encodes the source phrase into one or several fixed-size vectors,
and a decoder, which decodes the target phrase based on the results of the encoder.

\subsection{Encoder}
In contrast to the neural machine translation framework, 
our source language is not a proper language but instead a sequence of three variables making up a fact.
We propose an encoder sub-model, which encodes each atom of the fact into an embedding. Each atom
$\{s,r, o \}$, may stand for subject, relationship and object, respectively,  of a fact $F= (s, r,
o)$ is represented as a 1-of-$K$ vector $x_{\text{atom}}$, whose embedding is obtained as
$e_{\text{atom}}=E_{\text{in}} x_{\text{atom}}$, where $E_{\text{in}} \in \mathbb{R}^{D_{\text{Enc}} \times K}$ is the embedding matrix of the input vocabulary and $K$ is the size of that vocabulary. 
The encoder transforms this embedding into $\text{Enc}(F)_{\text{atom}}$ $\in \mathbb{R}^{H_{\text{Dec}}}$ 
as $\text{Enc}(F)_{\text{atom}} = W_{\text{Enc}} e_{\text{atom}}$, where $W_{\text{Enc}}$ $\in \mathbb{R}^{H_{\text{Dec}} \times D_{\text{Enc}}}$.

This embedding matrix, $E_{in}$, could be another parameter of the model to be learned, however, as discussed later (see Section \ref{modelSourceLanguage}), 
we have learned it separately and beforehand with \textit{TransE}~\cite{bordes2013translating}, 
a model aimed at modeling this kind of multi-relational data. We fix it and do not allow the encoder to tune it during training.

We call \textit{fact embedding} $\text{Enc}(F) \in \mathbb{R}^{3H_{\text{Dec}}}$ the concatenation $[ \text{Enc}(F)_{s}$, $\text{Enc}(F)_{r}$, $\text{Enc}(F)_{o}]$ of the atom embeddings, which is the input for the next module.

%

\subsection{Decoder} 
\begin{figure}
\centering
\includegraphics[natwidth=500px,natheight=753px,scale=0.31]{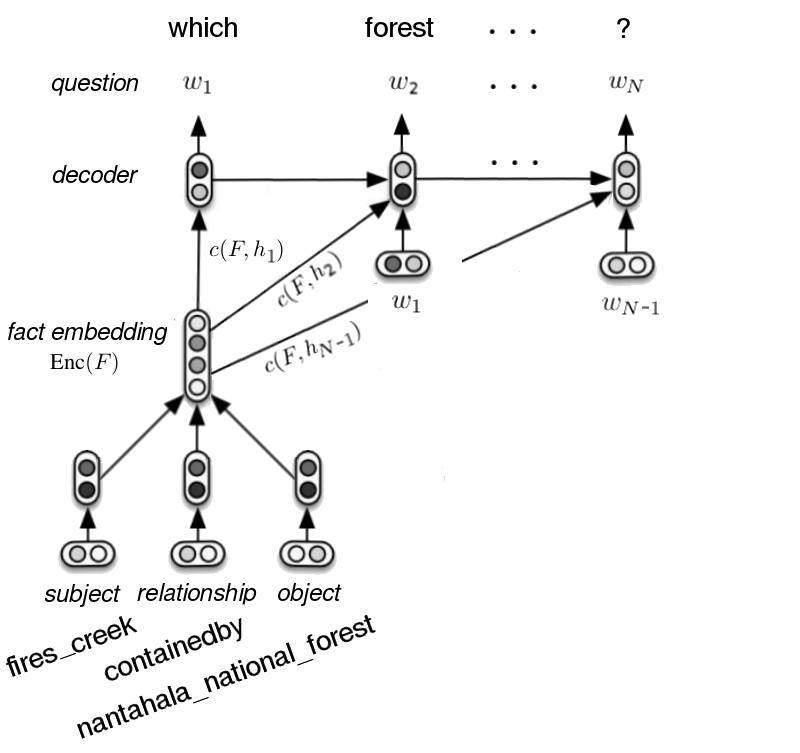}
\caption{\label{qa_model} The computational graph of the question-generation model,
where $\text{Enc}(F)$ is the fact embedding produced by the encoder model, and $c(F, h_{n-1})$ for $n=1,\dots,N$ is the fact representation weighed according to the attention-mechanism, which depends on both the fact $F$ and the previous hidden state of the decoder RNN $h_{n-1}$ .
For the sake of simplicity, the attention-mechanism
is not shown explicitly.
}
\end{figure}

For the decoder, we use a GRU recurrent neural network (RNN)~\cite{Cho-et-al-EMNLP2014} with an attention-mechanism~\cite{bahdanau2014neural} on the encoder representation to generate the associated question $Q$ to that fact $F$. 
Recently, it has been shown that the GRU RNN performs equally well across a range of tasks compared to other RNN architectures, such as the LSTM RNN \cite{greff2015lstm}.
The hidden state of the decoder RNN is computed at each time step $n$  as:
\begin{align}
& g^r_n = \sigma (W_r E_{out} w_{n-1}  + C_r c(F, h_{n-1}) + U_r h_{n-1}) \\
& g^u_n = \sigma (W_u E_{out} w_{n-1}  + C_u c(F, h_{n-1}) + U_u h_{n-1}) \\
& \tilde{h} = \text{tanh}(W E_{out} w_{n-1} + C c(F, h_{n-1}) \\ \nonumber
& \qquad \qquad + U (g^r_n \circ h_{n-1})) \\
& h_n =  g^u_n \circ h_{n-1} + (1 - g^u_n) \circ \tilde{h},
\end{align}
where $\sigma$ is the sigmoid function, s.t.~$\sigma(x) \in [0, 1]$, and the circle, $\circ$, represents element-wise multiplication. The initial state $h_0$ of this RNN is given by the output of a feedforward neural network fed with the fact embedding. 
The product $E_{out} w_n \in \mathbb{R}^{D_{\text{Dec}}}$ is the decoder embedding vector corresponding to the word $w_n$ (coded as a 1-of-$V$ vector, with $V$ being the size of the output vocabulary),
the variables $U_r, U_u, U, C_r, C_u, C \in \mathbb{R}^{H_{\text{Dec}} \times H_{\text{Dec}}}$, $W_r, W_u, W \in \mathbb{R}^{H_{\text{Dec}} \times D_{\text{Dec}}}$ are the parameters of the GRU and $c(F, h_{n-1})$ is the context vector (defined below Eq.~\ref{eq:c}).
The vector $g^r$ is called the \textit{reset gate}, 
$g^u$ as the \textit{update gate} 
and $\tilde{h}$ the \textit{candidate activation}.
By adjusting $g^r$ and $g^u$ appropriately, the model is able to create linear 
\textit{skip-connections} between distant hidden states, which in turn makes 
the credit assignment problem easier and the gradient signal stronger to
earlier hidden states. 
Then, at each time step $n$ the set of probabilities of word tokens is given by applying a softmax layer over $V_{o} tanh(V_{h}h_{n} + V_{w}E_{out} w_{n-1} + V_{c} c(F, h_{n-1}))$, where $V_{o} \in \mathbb{R}^{V \times H_{\text{Dec}}}$, $V_{h}, V_{c} \in \mathbb{R}^{H_{\text{Dec}} \times H_{\text{Dec}}}$ and $V_{w} \in \mathbb{R}^{H_{\text{Dec}} \times D_{\text{Dec}}}$.
Lastly, the function $c(F, h_{n-1})$ is computed using an attention-mechanism:




\begin{align}
& c(F, h_{n-1}) = \alpha_{\text{s},n-1}\text{Enc}(F)_{\text{s}} + \alpha_{\text{r},n-1}\text{Enc}(F)_{\text{r}} \nonumber \\
& \qquad + \alpha_{\text{o},n-1}\text{Enc}(F)_{\text{o}},
\label{eq:c}
\end{align}
where $\alpha_{\text{s},n-1}, \alpha_{\text{r},n-1}, \alpha_{\text{r},n-1}$ are real-valued scalars, which weigh the contribution of the subject, relationship and object representations. They correspond to the \textit{attention} of the model, and are computed by applying a one-layer neural network with tanh-activation function on the encoder representations of the fact, $\text{Enc}(F)$, and the previous hidden state of the RNN, $h_{n-1}$, followed by the sigmoid function to restrict the attention values to be between zero and one.
The need for the attention-mechanism is motivated by the intuition that the model needs to attend to the subject only once during the generation process while attending to the relationship at all other times during the generation process.
The model is illustrated in Figure \ref{qa_model}.




\subsection{Modeling the Source Language}
\label{modelSourceLanguage}
A particular problem with the model presented above is related to the embeddings for the entities, relationships and tokens,
which all have to be learned in one way or another.
If we learn these naively on the SimpleQuestions training set,
the model will perform poorly when it encounters previously unseen entities, relationships or tokens.
Furthermore, the multi-relational graph defined by the facts in SimpleQuestions is extremely sparse, 
i.e.\@ each node has very few edges to other nodes, as can be expected due to high ratio of unique entities over number of examples.
Therefore, even for many of the entities in SimpleQuestions, 
the model may perform poorly if the embedding is learned solely based on the SimpleQuestions dataset alone.

On the source side, we can resolve this issue by initializing the subject, relationship and object embeddings to those learned by applying multi-relational embedding-based models to the knowledge base.
Multi-relational embedding-based models~\cite{bordes-aaai-2011} 
have recently become popular to learn distributed vector embeddings for knowledge bases,
and have shown to scale well and yield good performance.
Due to its simplicity and good performance, we choose to use TransE~\cite{bordes2013translating} to learn such
embeddings. 
TransE is a translation-based model, whose energy function is trained to output low values when the fact expresses true information, i.e.\@ a fact which exists in the knowledge base, and otherwise high values. Formally, the energy function is defined as $f(s,r,o) = ||e_s + e_r - e_o ||_2$, where $e_s$, $e_r$ and $e_o$ are the real-valued embedding vectors for the subject, relationship and object of a fact.
Further details are given by Bordes et al.~\shortcite{bordes2013translating}.

 Embeddings for entities with few connections are easy to learn, 
yet the quality of these embeddings depends on how inter-connected they are. 
In the extreme case where the subject and object of a triple only appears once in the dataset, 
the learned embeddings of the subject and object will be semantically meaningless. 
This happens very often in SimpleQuestions, since only around 5$\%$ of the entities have more than 2 connections in the graph. 
Thus, by applying TransE directly over this set of triples, we would eventually end up with a layout of entities that does 
not contain clusters of semantically close concepts. In order to guarantee an effective semantic representation of the embeddings, 
we have to learn them together with additional triples extracted from the whole Freebase graph to complement the 
SimpleQuestions graph with relevant information for this task.

\begin{table*}
\centering
\small
\begin{tabular}{| c | c | c | c |}
\hline
{\bf Closest neighbors to} & Warner Bros. Entertainment & Manchester & hindi language \\\hline
\multirow{4}{*}{SQ} & Billy Gibbons & Ricky Anane & nepali indian\\
 & Jenny Lewis & Lee Dixon & Naseeb \\
 & Lies of Love & Jerri Bryne & Ghar Ek Mandir\\
 & Swordfish & Greg Wood & standard chinese \\
\hline
\multirow{4}{*}{SQ + FB} &  Paramount Pictures & Oxford & dutch language\\
& Sony Pictures Entertainment & Sale & italian language\\
& Electronic Arts & Liverpool & danish language\\
& CBS & Guildford & bengali language\\
\hline
\end{tabular}
\caption{Examples of differences in the local structure of the vector space embeddings when adding more FB facts}\label{tab:transEemb}
\end{table*}

We need a coarse representation for the entities contained in SimpleQuestions, 
capturing the \textit{basic} information, like the profession or nationality, the annotators tend to use when phrasing the questions, and accordingly we have ensured the embeddings contain this information by taking triples coming from the Freebase graph\footnote{Extracted from one of the latest Freebase dumps (downloaded by mid-August 2015) \url{https://developers.google.com/freebase/data}} regarding:
\begin{enumerate}
\item Category information: given by the \textit{type/instance} relationship,
this ensures that all the entities of the same semantic category are close to each other. 
Although one might think that the expected category of the subject/object could be inferred directly from the relationship, there are fine-grained differences in the expected types that be extracted only directly by observing this category information.
\item Geographical information: 
sometimes the annotators have included information about nationality 
(e.g. \textit{Which French president$\dots$?}) or location (e.g. \textit{Where in Germany$\dots$}?) of the subject and/or object.
This information is given by the relationships \textit{person/nationality} and \textit{location/contained$\_$by}.
By including these facts in the learning, we ensure the existence of a fine-grained layout of the embeddings regarding this information within a same category.
\item Gender: similarly, sometimes annotators have included information about gender (e.g. \textit{Which male audio engineer$\dots$}?). 
This information is given by the relationship \textit{person/gender}.
\end{enumerate}

To this end, we have included more than $300,000$ facts from Freebase in addition to the facts in SimpleQuestions for training. 
Table \ref{tab:transEemb} shows the differences in the embeddings before and after 
adding additional facts for training the TransE representations.



\subsection{Generating Questions}
To resolve the problem of data sparsity and previously unseen words on the target side,
we draw inspiration from the placeholders proposed 
for handling rare words in neural machine translation 
by Luong et al.~\shortcite{luong2015addressing}.
For every question and answer pair, we search for words in the question which overlap with words in the subject string of the fact.\footnote{We use the tool difflib: \url{https://docs.python.org/2/library/difflib.html}}
We heuristically estimate the sequence of most likely words in the question, which correspond to the subject string.
These words are then replaced by the placeholder token \textit{\textless placeholder\textgreater}. 
For example, given the fact \{fires\_creek, contained\_by, nantahala\_national\_forest\}
the original question \textit{Which forest is Fires Creek in?} is transformed into the question \textit{Which forest is \textless placeholder\textgreater  in?}.
The model is trained on these modified questions,
which means that model only has to learn decoder embeddings for tokens which are not in the subject string.
At test time, after outputting a question, all placeholder tokens are replaced by the subject string and then the outputs are evaluated.
We call this the Single-Placeholder (SP) model. The main difference with respect to that of Luong
et al.~\shortcite{luong2015addressing} is that we do not use placeholder tokens in the input language, because then the entities and relationships in the input would not be able to transmit semantic (e.g.\@ topical) information to the decoder. 
If we had included placeholder tokens in the input language, the model would not be able to generate informative words regarding the subject in the question (e.g.\@ it would be impossible for the model to learn that the subject \textit{Paris} may be accompanied by the words \textit{French city} when generating a question, because it would not see \textit{Paris} but only a placeholder token).

A single placeholder token for all question types could unnecessarily limit the model.
We therefore also experiment with another model, called the Multi-Placeholder (MP) model,
which uses $60$ different placeholder tokens such that the placeholder for a given question
is chosen based on the subject category extracted from the relationship (e.g. contained\_by is classified in the category location, and so the transformed question would be \textit{Which forest is \ \textless location placeholder\textgreater \ in?}).
This could make it easier for the model to learn to phrase questions about a diverse set of entities,
but it also introduces additional parameters, since there are now $60$ placeholder embeddings to be learned,
and therefore the model may suffer from overfitting.
This way of addressing the sparsity in the output reduces the vocabulary size to less than 7000 words.

\subsection{Template-based Baseline}
To compare our neural network models, we propose a (non-parametric) template-based baseline model, 
which makes use of the entire training set when generating a question.
The baseline operates on questions modified with the placeholder as in the preceding section. 
Given a fact $F$ as input, the baseline picks a candidate fact $F_c$ in the training set at uniformly random,
where $F_c$ has the same relationship as $F$. 
Then the baseline considers the questions corresponding to $F_c$ and as in the SP model, in the final step the placeholder token in the question is replaced by the subject string of the fact $F$. 

\section{Experiments}
\subsection{Training Procedure}

All models were implemented using Theano \cite{2016arXiv160502688short}.
To train the neural network models, we optimized the log-likelihood using the
first-order gradient-based optimization algorithm Adam \cite{kingma2014adam}.
To decide when to stop training we used early stopping with patience \cite{bengio2012practical} on the METEOR score obtained for the validation set.
In all experiments, we use the default split of the SimpleQuestions dataset into training, validation and test sets.

We trained TransE embeddings with embedding dimensionality $200$ for each subject, relationship and object.
Based on preliminary experiments, for all neural network models
we fixed the learning rate to $0.00025$ and clipped parameter gradients with norms larger than $0.1$.
We further fixed the embedding dimensionality of words to be $200$,
and the hidden state of the decoder RNN to have dimensionality $600$.


\subsection{Evaluation}

To investigate the performance of our models, we make use of both automatic evaluation metrics and human evaluators.

\begin{table*}[t]
  \small
  \centering
    \begin{tabular}{| c | c | c | c | c | c |}
    \hline
    \textbf{Model} & \textbf{BLEU} & \textbf{METEOR} & \textbf{Emb. Greedy} \\ \hline
    Baseline & 31.36 & 33.12 & 74.02 \\ \hline
    SP Triples & 33.27 & 35.07 & 76.72 \\ \hline
    MP Triples & 32.76 & 34.97 & 76.70 \\ \hline
    SP Triples TransE++ & \textbf{33.32} & \textbf{35.38} & 76.78 \\ \hline
    MP Triples TransE++ & 33.28 & 35.29 & \textbf{77.01} \\ \hline
    \end{tabular}
    \caption{Test performance for all models w.r.t. BLEU, METEOR and Emb.\@ Greedy performance metrics, where \textit{SP} indicates models with a single placeholder and \textit{MP} models with multiple placeholders. \textit{TransE++} indicates models where the TransE embeddings have been pretrained on a larger set of triples. The best performance on each metric is marked in bold font.}
\label{table:automatic_test_results}
\end{table*}


\subsubsection{Automatic Evaluation Metrics}
BLEU~\cite{papineni2002bleu} and METEOR~\cite{banerjee2005meteor} are two widely used evaluation
metrics in statistical machine translation and automatic image-caption generation \cite{chen2015microsoft}.
Similar to statistical machine translation, where a phrase in the source language is mapped to a phrase in the target language,
in this task a KB fact is mapped to a natural language question. 
Both tasks are highly constrained, e.g.\@ the set of valid outputs is limited.
This is true in particular for short phrases, such as one sentence questions.
Furthermore, in both tasks, 
the majority of valid outputs are paraphrases of each other,
which BLEU and METEOR have been designed to capture.
We therefore believe that BLEU and METEOR constitute reasonable performance metrics for evaluating the generated questions.



Although we believe that METEOR and BLEU are reasonable evaluation metrics, 
they may have not recognize certain paraphrases, in particular paraphrases of entities.
We therefore also make use of a sentence similarity metric,
as proposed by Rus and Lintean \shortcite{rus2012comparison},
which we will denote \textit{Embedding Greedy} (Emb.\@ Greedy).
The metric makes use of a word similarity score,
which in our experiments is the cosine similarity between two Word2Vec word embeddings \cite{mikolov2013distributed}.\footnote{We use the Word2Vec embeddings pretrained on the Google News Corpus: \url{https://code.google.com/p/word2vec/}.} 
The metric finds a (non-exclusive) alignment between words in the two questions,
which maximizes the similarity between aligned words,
and computes the sentence similarity as the mean over the word similarities between aligned words.

\begin{table*}[htpb]
\renewcommand\arraystretch{1.1}
\centering
  \small
  \centering
    \begin{tabular}{| c | c | c | c |}
    \hline
    \textbf{Fact} & \textbf{Human} & \textbf{Baseline} & \textbf{MP Triples TransE++} \\ \hline
     \parbox[t]{3cm}{\centering bayuvi dupki \\ -- contained by -- \\ europe} & \parbox[t]{3cm}{where is bayuvi dupki?} & \parbox[t]{3cm}{what state is the city of bayuvi dupki located in?} & \parbox[t]{3.5cm}{what continent is bayuvi dupki in?}
 \\ \hline
     \parbox[t]{3cm}{\centering illinois \\ -- contains -- \\ ludlow township} & \parbox[t]{3cm}{what is in illinois?} & \parbox[t]{3cm}{what is a tributary found in illinois?} & \parbox[t]{3.5cm}{what is the name of a place within illinois?} \\ \hline
     \parbox[t]{3cm}{\centering neo contra \\ -- publisher -- \\ konami} & \parbox[t]{3cm}{who published \\ neo contra?} & \parbox[t]{3cm}{which company published the game neo contra?} & \parbox[t]{3.5cm}{who is the publisher for the computer videogame neo contra?} \\[\dimexpr+\normalbaselineskip+10pt] \hline 
     \parbox[t]{3cm}{\centering pop music \\ -- artists -- \\ nikki flores} & \parbox[t]{3cm}{what artist is known for pop music?} & \parbox[t]{3.5cm}{An example of pop music is what artist?} & \parbox[t]{3cm}{who's an american singer that plays pop music?} \\[\dimexpr+\normalbaselineskip+10pt] \hline
    \end{tabular}
    \caption{Test examples and corresponding questions using the template-based baseline and MP Triples TransE++ model. For more examples, please refer to the supplementary material.}
\label{table:examples}
\end{table*}

The results are shown in Table \ref{table:automatic_test_results}. 
Example questions produced by the model with multiple placeholders are shown in Table \ref{table:examples}.
The neural network models outperform the template-based baseline by a clear margin across all metrics.
Template-based baseline is already a relatively strong model, because it makes use of a separate template for each relationship.
Given enough training data this suggests that neural networks are generally better at the question generation task compared to hand-crafted template-based procedures,
and therefore that they may be useful for generating question answering corpora. 
Furthermore, it appears that the best performing models are the models where TransE are trained on the largest set of triples (TransE++).  This set contains, apart from the supporting triples described in Section \ref{modelSourceLanguage}, triples involving entities which are highly connected to the entities found in the SimpleQuestions facts. 
In total, around 30 millions of facts, which have been used to generate the 30M Factoid Question-Answer Corpus.
Lastly, it is not clear whether the model with a single placeholder or the model with multiple placeholders performs best.
This motivates the following human evaluation study.

\subsubsection{Human Evaluation Study}

\begin{table*}[htpb]
  \small
  \centering
    \begin{tabular}{| c | c | c | c | c |}
    \hline
    \textbf{Model A} & \textbf{Model B} & \textbf{Model A Preference (\%)} & \textbf{Model B Preference (\%)} & \textbf{Fleiss' kappa} \\ \hline
    Human & Baseline & $^*\mathbf{56.329 \pm 5.469}$ & $34.177 \pm 5.230$ & $0.242$ \\ \hline
    Baseline & MP Triples TransE++ & $32.484 \pm 5.180$ & $^*\mathbf{60.828 \pm 5.399}$ & $0.234$  \\ \hline
    Human & MP Triples TransE++ & $38.652 \pm 5.684$ & $\mathbf{51.418 \pm 5.833}$ & $0.182$ \\ \hline
    \end{tabular}
    \caption{
Pairwise human evaluation preferences computed across evaluators with $95\%$ confidence intervals.
The preferred model in each experiment is marked in bold font. An asterisk next to the preferred model indicates a statistically significance likelihood-ratio test, which shows that the model is preferred in at least half of the presented examples with $95\%$ confidence.
The name \textit{MP Triples TransE++} indicates the model with multiple placeholders and TransE embeddings pretrained on a larger set of triples.
The last column shows the Fleiss' kappa averaged across batches (HITs) with different evaluators and questions.}
\label{table:human_test_results}
\end{table*}

We carry out pairwise preference experiments on Amazon Mechanical Turk.

Initially, we considered carrying out separate experiments for measuring relevancy and fluency respectively,
since this is common practice in machine translation.
However, the relevancy of a question is determined solely by a single factor,
i.e.\@ the relationship, since by construction the subject is always in the question.
Measuring relevancy is therefore not very useful in our task.
To verify this we carried out an internal pairwise preference experiment with human subjects,
who were repeatedly shown a fact and two questions and asked to select the most relevant question.
We found that $93\%$ of the questions generated by the MP Triples TransE++ model were either judged better or at least as good as the human generated questions w.r.t.\@ relevancy.
The remaining $7\%$ questions of the MP Triples TransE++ model questions were also judged relevant questions,
although less so compared to the human generated questions.
In the next experiment, we therefore measure the holistic quality of the questions.

We setup experiments comparing:
Human-Baseline (human and baseline questions), 
Human-MP (human and MP Triples TransE++ questions) and Baseline-MP (baseline and MP Triples TransE++ questions).
We show human evaluators a fact along with two questions, one question from each model for the corresponding fact,
and ask the them to choose the question which is most relevant to the fact and most natural.
The human evaluator also has the option of not choosing either question.
This is important if both questions are equally good 
or if neither of the questions make sense.
At the beginning of each experiment, we show the human evaluators two examples of statements and a corresponding pair of questions,
where we briefly explain the form of the statements and how questions relate to those statements.
Following the introductory examples, we present the facts and corresponding pair of questions one by one.
To avoid presentation bias, we randomly shuffle the order of the examples and the order in which questions are shown by each model.
During each experiment, we also show four check facts and corresponding check questions at random,
which any attentive human annotator should be able to answer easily.
We discard responses of human evaluators who fail any of these four checks.

The preference of each example is defined as the question which is preferred by the majority of the evaluators.
Examples where neither of the two questions are preferred by the majority of the evaluators,
i.e.\@ when there is an equal number of evaluators who prefer each question,
are assigned to a separate preference class called ``comparable''.\footnote{The probabilities for the ``comparable'' class in Table \ref{table:human_test_results} can be computed in each row as $100$ minus the third and fourth column in the table.}

The results are shown in Table \ref{table:human_test_results}.
In total, $3,810$ preferences were recorded by $63$ independent human evaluators.
The questions produced by each model model pair were evaluated in $5$ batches (HITs). 
Each human evaluated $44$-$75$ examples (facts and corresponding question pairs) in each batch and
each example was evaluated by $3$-$5$ evaluators.
In agreement with the automatic evaluation metrics, 
the human evaluators strongly prefer either the human or the neural network model over the template-based baseline.
Furthermore, it appears that humans cannot distinguish between the human-generated questions and the neural network questions,
on average showing a preference towards the later over the former ones.
We hypothesize this is because our model penalizes uncommon and unnatural ways to frame questions\footnote{Upon further inspection, we believe that some questions of the SimpleQuestions dataset have been produced by non-native English speakers.}
and, sometimes, includes specific information about the target object that the humans do not (see last example of Table \ref{table:examples}).
This confirms our earlier assertion, that the neural network questions can be used for building question answering systems.

\section{Conclusion}
We propose new neural network models for mapping knowledge base facts into corresponding natural language questions.
The neural networks combine ideas from recent neural network architectures for statistical machine translation,
as well as multi-relational knowledge base embeddings for overcoming sparsity issues
and placeholder techniques for handling rare words.
The produced question and answer pairs are evaluated using automatic evaluation metrics,
including BLEU, METEOR and sentence similarity,
and are found to outperform a template-based baseline model.
When evaluated by untrained human subjects, 
the question and answer pairs produced by our best performing neural network 
appear comparable in quality to the real human-generated questions.
Finally, we use our best performing neural network model to generate a corpus of $30$M question and answer pairs,
which we hope will enable future researchers to improve their question answering systems.

\section*{Acknowledgments}

The authors acknowledge IBM Research, NSERC, Canada
Research Chairs and CIFAR for funding.
The authors thank Yang Yu, Bing Xiang, Bowen Zhou and Gerald Tesauro
for constructive feedback,
and  Antoine Bordes, Nicolas Usunier, Sumit Chopra and Jason Weston
for providing the SimpleQuestions dataset.
This research was enabled in part by support provided by 
Calcul Québec (\url{www.calculquebec.ca}) and Compute Canada (\url{www.computecanada.ca}).


\bibliography{acl2016}

\begin{thebibliography}{}

\bibitem[\protect\citename{Ali \bgroup et al.\egroup }2010]{ali2010automation}
Husam Ali, Yllias Chali, and Sadid~A Hasan.
\newblock 2010.
\newblock Automation of question generation from sentences.
\newblock In {\em Proceedings of QG2010: The Third Workshop on Question
  Generation}, pages 58--67.

\bibitem[\protect\citename{Bahdanau \bgroup et al.\egroup
  }2015]{bahdanau2014neural}
Dzmitry Bahdanau, Kyunghyun Cho, and Yoshua Bengio.
\newblock 2015.
\newblock Neural machine translation by jointly learning to align and
  translate.
\newblock In {\em International Conference on Learning Representations}.

\bibitem[\protect\citename{Banerjee and Lavie}2005]{banerjee2005meteor}
Satanjeev Banerjee and Alon Lavie.
\newblock 2005.
\newblock {METEOR}: An automatic metric for mt evaluation with improved
  correlation with human judgments.
\newblock In {\em Proceedings of the ACL workshop on intrinsic and extrinsic
  evaluation measures for machine translation and/or summarization}, volume~29,
  pages 65--72.

\bibitem[\protect\citename{Bengio}2012]{bengio2012practical}
Yoshua Bengio.
\newblock 2012.
\newblock Practical recommendations for gradient-based training of deep
  architectures.
\newblock In {\em Neural Networks: Tricks of the Trade}, pages 437--478.
  Springer.

\bibitem[\protect\citename{Berant and Liang}2014]{berant2014semantic}
Jonathan Berant and Percy Liang.
\newblock 2014.
\newblock Semantic parsing via paraphrasing.
\newblock In {\em Proceedings of ACL}, volume~7, pages 1415--1425.

\bibitem[\protect\citename{Bollacker \bgroup et al.\egroup
  }2008]{bollacker2008freebase}
Kurt Bollacker, Colin Evans, Praveen Paritosh, Tim Sturge, and Jamie Taylor.
\newblock 2008.
\newblock Freebase: a collaboratively created graph database for structuring
  human knowledge.
\newblock In {\em Proceedings of the 2008 ACM SIGMOD international conference
  on Management of data}, pages 1247--1250.

\bibitem[\protect\citename{Bordes \bgroup et al.\egroup
  }2011]{bordes-aaai-2011}
Antoine Bordes, Jason Weston, Ronan Collobert, and Yoshua Bengio.
\newblock 2011.
\newblock Learning structured embeddings of knowledge bases.
\newblock In {\em AAAI 2011}.

\bibitem[\protect\citename{Bordes \bgroup et al.\egroup
  }2013]{bordes2013translating}
Antoine Bordes, Nicolas Usunier, Alberto Garcia-Duran, Jason Weston, and Oksana
  Yakhnenko.
\newblock 2013.
\newblock Translating embeddings for modeling multi-relational data.
\newblock In {\em Advances in Neural Information Processing Systems}, pages
  2787--2795.

\bibitem[\protect\citename{Bordes \bgroup et al.\egroup }2014]{bordes2014QA}
Antoine Bordes, Jason Weston, and Nicolas Usunier.
\newblock 2014.
\newblock Open question answering with weakly supervised embedding models.
\newblock In {\em Machine Learning and Knowledge Discovery in Databases -
  European Conference, (ECML PKDD)}, pages 165--180.

\bibitem[\protect\citename{Bordes \bgroup et al.\egroup }2015]{bordes2015large}
Antoine Bordes, Nicolas Usunier, Sumit Chopra, and Jason Weston.
\newblock 2015.
\newblock Large-scale simple question answering with memory networks.
\newblock {\em arXiv preprint arXiv:1506.02075}.

\bibitem[\protect\citename{Chen \bgroup et al.\egroup }2009]{chen2009aist}
Wei Chen, Gregory Aist, and Jack Mostow.
\newblock 2009.
\newblock Generating questions automatically from informational text.
\newblock In {\em Proceedings of the 2nd Workshop on Question Generation (AIED
  2009)}, pages 17--24.

\bibitem[\protect\citename{Chen \bgroup et al.\egroup }2015]{chen2015microsoft}
Xinlei Chen, Hao Fang, Tsung-Yi Lin, Ramakrishna Vedantam, Saurabh Gupta, Piotr
  Dollar, and C.~Lawrence Zitnick.
\newblock 2015.
\newblock Microsoft {COCO} captions: Data collection and evaluation server.
\newblock {\em arXiv preprint arXiv:1504.00325}.

\bibitem[\protect\citename{Cho \bgroup et al.\egroup
  }2014]{Cho-et-al-EMNLP2014}
Kyunghyun Cho, Bart van Merrienboer, Caglar Gulcehre, Fethi Bougares, Holger
  Schwenk, and Yoshua Bengio.
\newblock 2014.
\newblock Learning phrase representations using {RNN} encoder-decoder for
  statistical machine translation.
\newblock In {\em Proceedings of the Conference on Empirical Methods in Natural
  Language Processing}, pages 1724--1734.

\bibitem[\protect\citename{Curto \bgroup et al.\egroup
  }2012]{curto2012question}
Sergio Curto, A~Mendes, and Luisa Coheur.
\newblock 2012.
\newblock Question generation based on lexico-syntactic patterns learned from
  the web.
\newblock {\em Dialogue and Discourse}, 3(2):147--175.

\bibitem[\protect\citename{Duma and Klein}2013]{duma2013generating}
Daniel Duma and Ewan Klein.
\newblock 2013.
\newblock Generating natural language from linked data: Unsupervised template
  extraction.
\newblock {\em ACL}, pages 83--94.

\bibitem[\protect\citename{Dumais \bgroup et al.\egroup }2002]{dumais2002web}
Susan Dumais, Michele Banko, Eric Brill, Jimmy Lin, and Andrew Ng.
\newblock 2002.
\newblock Web question answering: Is more always better?
\newblock In {\em Proceedings of the 25th annual international ACM SIGIR
  conference on Research and development in information retrieval}, pages
  291--298.

\bibitem[\protect\citename{Graesser \bgroup et al.\egroup
  }1992]{graesser1992quest}
Arthur~C Graesser, Sallie~E Gordon, and Lawrence~E Brainerd.
\newblock 1992.
\newblock {QUEST}: A model of question answering.
\newblock {\em Computers and Mathematics with Applications}, 23(6):733--745.

\bibitem[\protect\citename{Greff \bgroup et al.\egroup }2015]{greff2015lstm}
Klaus Greff, Rupesh~Kumar Srivastava, Jan Koutn{\'\i}k, Bas~R Steunebrink, and
  J{\"u}rgen Schmidhuber.
\newblock 2015.
\newblock {LSTM}: A search space odyssey.
\newblock {\em arXiv preprint arXiv:1503.04069}.

\bibitem[\protect\citename{Kalady \bgroup et al.\egroup
  }2010]{kalady2010natural}
Saidalavi Kalady, Ajeesh Elikkottil, and Rajarshi Das.
\newblock 2010.
\newblock Natural language question generation using syntax and keywords.
\newblock In {\em Proceedings of QG2010: The Third Workshop on Question
  Generation}, pages 1--10. questiongeneration. org.

\bibitem[\protect\citename{Kingma and Ba}2015]{kingma2014adam}
Diederik Kingma and Jimmy Ba.
\newblock 2015.
\newblock Adam: A method for stochastic optimization.
\newblock In {\em The International Conference on Learning Representations}.

\bibitem[\protect\citename{Lenat and Guha}1989]{lenat1989building}
Douglas~B. Lenat and Ramanathan~V. Guha.
\newblock 1989.
\newblock {\em Building large knowledge-based systems; representation and
  inference in the Cyc project}.
\newblock Addison-Wesley Longman Publishing Co., Inc.

\bibitem[\protect\citename{Lopez \bgroup et al.\egroup
  }2011]{lopez2011question}
Vanessa Lopez, Victoria Uren, Marta Sabou, and Enrico Motta.
\newblock 2011.
\newblock Is question answering fit for the semantic web? a survey.
\newblock {\em Semantic Web}, 2(2):125--155.

\bibitem[\protect\citename{Luong \bgroup et al.\egroup
  }2015]{luong2015addressing}
Minh-Thang Luong, Ilya Sutskever, Quoc~V Le, Oriol Vinyals, and Wojciech
  Zaremba.
\newblock 2015.
\newblock Addressing the rare word problem in neural machine translation.
\newblock In {\em Proceedings of ACL}, pages 11--19.

\bibitem[\protect\citename{Mannem \bgroup et al.\egroup
  }2010]{mannem2010question}
Prashanth Mannem, Rashmi Prasad, and Aravind Joshi.
\newblock 2010.
\newblock Question generation from paragraphs at upenn: Qgstec system
  description.
\newblock In {\em Proceedings of QG2010: The Third Workshop on Question
  Generation}, pages 84--91.

\bibitem[\protect\citename{Mikolov \bgroup et al.\egroup
  }2013]{mikolov2013distributed}
Tomas Mikolov, Ilya Sutskever, Kai Chen, Greg~S Corrado, and Jeff Dean.
\newblock 2013.
\newblock Distributed representations of words and phrases and their
  compositionality.
\newblock In {\em Advances in Neural Information Processing Systems}, pages
  3111--3119.

\bibitem[\protect\citename{Olney \bgroup et al.\egroup
  }2012]{olney2012question}
Andrew~M Olney, Arthur~C Graesser, and Natalie~K Person.
\newblock 2012.
\newblock Question generation from concept maps.
\newblock {\em Dialogue and Discourse}, 3(2):75--99.

\bibitem[\protect\citename{Papineni \bgroup et al.\egroup
  }2002]{papineni2002bleu}
Kishore Papineni, Salim Roukos, Todd Ward, and Wei-Jing Zhu.
\newblock 2002.
\newblock Bleu: a method for automatic evaluation of machine translation.
\newblock In {\em Proceedings of the 40th annual meeting on ACL}, pages
  311--318.

\bibitem[\protect\citename{Rus and Lintean}2012]{rus2012comparison}
Vasile Rus and Mihai Lintean.
\newblock 2012.
\newblock A comparison of greedy and optimal assessment of natural language
  student input using word-to-word similarity metrics.
\newblock In {\em Proceedings of the Seventh Workshop on Building Educational
  Applications Using NLP, NAACL}, pages 157--162.

\bibitem[\protect\citename{Rus \bgroup et al.\egroup }2010]{rus2010first}
Vasile Rus, Brendan Wyse, Paul Piwek, Mihai Lintean, Svetlana Stoyanchev, and
  Cristian Moldovan.
\newblock 2010.
\newblock The first question generation shared task evaluation challenge.
\newblock In {\em Proceedings of the 6th International Natural Language
  Generation Conference}, pages 251--257.

\bibitem[\protect\citename{Sutskever \bgroup et al.\egroup
  }2014]{sutskever2014sequence}
Ilya Sutskever, Oriol Vinyals, and Quoc~V. Le.
\newblock 2014.
\newblock Sequence to sequence learning with neural networks.
\newblock In {\em Advances in Neural Information Processing Systems}.

\bibitem[\protect\citename{{Theano Development
  Team}}2016]{2016arXiv160502688short}
{Theano Development Team}.
\newblock 2016.
\newblock {Theano: A {Python} framework for fast computation of mathematical
  expressions}.
\newblock {\em arXiv e-prints}, abs/1605.02688, May.

\bibitem[\protect\citename{Voorhees and Tice}2000]{voorhees2000overview}
Ellen~M Voorhees and DM~Tice.
\newblock 2000.
\newblock Overview of the trec-9 question answering track.
\newblock In {\em TREC}.

\bibitem[\protect\citename{Vrande{\v{c}}i{\'c} and
  Kr{\"o}tzsch}2014]{vrandevcic2014wikidata}
Denny Vrande{\v{c}}i{\'c} and Markus Kr{\"o}tzsch.
\newblock 2014.
\newblock Wikidata: a free collaborative knowledgebase.
\newblock {\em Communications of the ACM}, 57(10):78--85.

\bibitem[\protect\citename{Yao and Zhang}2010]{yao2010question}
Xuchen Yao and Yi~Zhang.
\newblock 2010.
\newblock Question generation with minimal recursion semantics.
\newblock In {\em Proceedings of QG2010: The Third Workshop on Question
  Generation}, pages 68--75.

\bibitem[\protect\citename{Yao \bgroup et al.\egroup }2012]{yao2012semantics}
Xuchen Yao, Gosse Bouma, and Yi~Zhang.
\newblock 2012.
\newblock Semantics-based question generation and implementation.
\newblock {\em Dialogue and Discourse}, 3(2):11--42.

\end{thebibliography}
\bibliographystyle{acl2016}

\appendix


\onecolumn

\section{Supplemental Material: Generated Questions}
\begin{table*}[htpb]
\renewcommand\arraystretch{1.1}
\centering
  \small
  \centering
    \begin{tabular}{| c | c | c | c |}
    \hline
    \textbf{Fact} & \textbf{Human} & \textbf{Baseline} & \textbf{MP Triples TransE++} \\ \hline
 \parbox[t]{3cm}{\centering make your mind \\ -- format -- \\ digital media}   &   \parbox[t]{3cm}{what is make your mind?} & \parbox[t]{3cm}{what was the format for the release make your mind?}   &  \parbox[t]{3.5cm}{what format is make your mind released?} \\ \hline 
 \parbox[t]{3cm}{\centering west air \\ -- hubs -- \\ san diego international airport}   &   \parbox[t]{3cm}{what west air's hub?} &      \parbox[t]{3cm}{west air flies out of which airport?}   &  \parbox[t]{3.5cm}{which airport is named after the hub?} \\ \hline
 \parbox[t]{4cm}{\centering fumihiko maki \\ -- structures designed -- \\ makuhari messe}   &   \parbox[t]{3cm}{fumihiko maki designed what structure?} &      \parbox[t]{3cm}{what park did fumihiko maki help design?}   &  \parbox[t]{3.5cm}{what's a structure designed by fumihiko maki?} \\ \hline
 \parbox[t]{3cm}{\centering sulfadoxine \\ -- used to treat  -- \\ falciparum malaria}   &   \parbox[t]{3cm}{what does sulfadoxine help treat?} & \parbox[t]{3cm}{what disease is sulfadoxine used to treat?}   &  \parbox[t]{3.5cm}{what is used to treat sulfadoxine?} \\ \hline 
 \parbox[t]{3cm}{\centering painting \\ -- artists  -- \\ adam willaerts}   &   \parbox[t]{3cm}{which artist specialize in painting?} & \parbox[t]{3cm}{whos an artist that specializes in painting?}   &  \parbox[t]{3.5cm}{what is the name of an artist that creates visual painting?} \\ \hline 
 \parbox[t]{3cm}{\centering africa \\ -- recorded by  -- \\ dwight trible}   &   \parbox[t]{3cm}{who released the recording africa?} & \parbox[t]{3cm}{what is the name if a british indie rock band that played the africa album?}   &  \parbox[t]{3.5cm}{who recorded africa?} \\ \hline 
 \parbox[t]{3cm}{\centering tom larkin \\ -- place musical career began  -- \\ wellington}   &   \parbox[t]{3cm}{where is tom larkin from?} & \parbox[t]{3cm}{where in the us is tom larkin originally from?}   &  \parbox[t]{3.5cm}{where does tom larkin originate from?} \\ \hline 
 \parbox[t]{3cm}{\centering film director \\ -- people with this profession -- \\ r. t. neason}   &   \parbox[t]{3cm}{name a famous film director?} &      \parbox[t]{3cm}{who was a film director?}   &  \parbox[t]{3.5cm}{name a person who works as a film director?} \\[\dimexpr+\normalbaselineskip+23pt] \hline
 \parbox[t]{4cm}{\centering abyss \\ -- powers or abilities -- \\ shapeshifting}   &   \parbox[t]{3cm}{what is abyss' power?} &      \parbox[t]{3cm}{what is abyss's power or ability?}   &  \parbox[t]{3.5cm}{what is abyss's special ability?} \\ \hline
 \parbox[t]{3cm}{\centering drowning \\ -- release  -- \\ laughter \& lust}   &   \parbox[t]{3cm}{which release is drowning on?} & \parbox[t]{3cm}{what is the release track on drowning?}   &  \parbox[t]{3.5cm}{which track is part of drowning?} \\ \hline 
 \parbox[t]{4cm}{\centering masse broberg \\ -- country of nationality  -- \\ sweden}   &   \parbox[t]{3cm}{where is masse broberg from?} & \parbox[t]{3cm}{what is masse broberg's country of nationality?}   &  \parbox[t]{3.5cm}{what country is masse broberg from?} \\[\dimexpr+\normalbaselineskip+5pt] \hline 
    \end{tabular}
    \caption{Test examples and corresponding questions using the template-based baseline and MP Triples TransE++.}
\label{table:appendix_examples}
\end{table*}


\begin{table*}[htpb]
\renewcommand\arraystretch{1.1}
\centering
  \small
  \centering
    \begin{tabular}{| c | c | c | c |}
    \hline
    \textbf{Fact} & \textbf{Human} & \textbf{Baseline} & \textbf{MP Triples TransE++} \\ \hline
 \parbox[t]{4cm}{\centering another day \\ -- region  -- \\ united states of america}   &   \parbox[t]{3cm}{where was another day released?} &      \parbox[t]{3cm}{which region was another day released in?}   &  \parbox[t]{3.5cm}{which country was another day released in?} \\[\dimexpr+\normalbaselineskip+10pt] \hline 
 \parbox[t]{3cm}{\centering presque isle \\ -- county  -- \\ aroostook county}   &   \parbox[t]{3cm}{what county contains presque isle?} &      \parbox[t]{3cm}{which county is presque isle located in?}   &  \parbox[t]{3.5cm}{what county is presque isle in?} \\ \hline 
 \parbox[t]{3cm}{\centering africa \\ -- release  -- \\ enjoy the mix}   &   \parbox[t]{3cm}{what was africa released on?} &      \parbox[t]{3cm}{what album includes africa?}   &  \parbox[t]{3.5cm}{what is the release track on africa?} \\ \hline 
 \parbox[t]{4cm}{\centering cherry \\ -- drugs with this flavor  -- \\ tussin expectorant for adults 100 syrup}   &   \parbox[t]{3cm}{name a cherry flavored drug?} &      \parbox[t]{3cm}{what is a cherry flavored drug?}   &  \parbox[t]{3.5cm}{what's a drug that cherry shaped like?} \\ \hline 
 \parbox[t]{3cm}{\centering jack tafari \\ -- notable types  -- \\ person}   &   \parbox[t]{3cm}{what organism is jack tafari?} &      \parbox[t]{3cm}{what is jack tafari known as?}   &  \parbox[t]{3.5cm}{what is jack tafari known as?} \\ \hline 
 \parbox[t]{3cm}{\centering rendsburg \\ -- people born here  -- \\ jost de jager}   &   \parbox[t]{3cm}{who was born in rendsburg?} &      \parbox[t]{3cm}{who was born in rendsburg?}   &  \parbox[t]{3.5cm}{who is a notable person that was born in rendsburg?} \\ \hline 
 \parbox[t]{3cm}{\centering cuveglio \\ -- contained by  -- \\ italy}   &   \parbox[t]{3cm}{what country is cuveglio in?} &      \parbox[t]{3cm}{where is the city of cuveglio located in?}   &  \parbox[t]{3.5cm}{what country is cuveglio located in?} \\ \hline 
 \parbox[t]{3cm}{\centering eduards veidenbaums \\ -- place of death  -- \\ russian empire}   &   \parbox[t]{3cm}{where was eduards veidenbaums deceased?} &      \parbox[t]{3cm}{which city did eduards veidenbaums die in?}   &  \parbox[t]{3.5cm}{where did eduards veidenbaums die?} \\ \hline 
 \parbox[t]{3cm}{\centering joe smith \\ -- place of death  -- \\ somme}   &   \parbox[t]{3cm}{where did joe smith die?} &      \parbox[t]{3cm}{where did joe smith die?}   &  \parbox[t]{3.5cm}{where did joe smith's life end?} \\[\dimexpr+\normalbaselineskip+10pt] \hline 
 \parbox[t]{3cm}{\centering electro-industrial \\ -- albums  -- \\ saw}   &   \parbox[t]{3cm}{what is an electro-industrial genre?} &      \parbox[t]{3cm}{which album is a electro-industrial album?}   &  \parbox[t]{3.5cm}{what is the name of a electro-industrial album?} \\[\dimexpr+\normalbaselineskip+10pt] \hline 
 \parbox[t]{3cm}{\centering in these arms \\ -- composer  -- \\ richie sambora}  &   \parbox[t]{3cm}{who composed in these arms?} &     \parbox[t]{3cm}{who is the creator of in these arms?}  &  \parbox[t]{3.5cm}{who composed in these arms?} \\[\dimexpr+\normalbaselineskip+10pt] \hline 
 \parbox[t]{4cm}{\centering baroque \\ -- associated artworks -- \\ the rainbow landscape}   &   \parbox[t]{3cm}{which artwork is in baroque?} &      \parbox[t]{3cm}{name an artwork associated with the baroque art period movement?}   &  \parbox[t]{3.5cm}{what is a piece of art with a baroque artwork?} \\ \hline
 \parbox[t]{3cm}{\centering 11664 kashiwagi \\ -- orbits -- \\ sun}   &   \parbox[t]{3cm}{what does 11664 kashiwagi orbit?} &      \parbox[t]{3cm}{which orbit has relationship with 11664 kashiwagi?}   &  \parbox[t]{3.5cm}{around which main star does 11664 kashiwagi gravitate?} \\ \hline
 \parbox[t]{3cm}{\centering harbord collegiate institute \\ -- notable types -- \\ school}   &   \parbox[t]{3cm}{what is harbord collegiate institute?} &      \parbox[t]{3cm}{what is harbord collegiate institute known for being?}   &  \parbox[t]{3.5cm}{what type of building is the harbord collegiate institute?} \\[\dimexpr+\normalbaselineskip+10pt] \hline
 \parbox[t]{3cm}{\centering milwaukee \\ -- neighborhoods -- \\ arlington heights}   &   \parbox[t]{3cm}{what neighborhood is found in milwaukee?} &      \parbox[t]{3cm}{what's a neighborhood in the portland milwaukee?}   &  \parbox[t]{3.5cm}{what is the name of a neighborhood in milwaukee florida?} \\ \hline
 \parbox[t]{3cm}{\centering cheryl hickey \\ -- profession -- \\ actor}   &   \parbox[t]{3cm}{what is cheryl hickey's profession?} &      \parbox[t]{3cm}{what is cheryl hickey?}   &  \parbox[t]{3.5cm}{what is cheryl hickey's profession in the entertainment industry?} \\ \hline
    \end{tabular}
    \caption{Test examples and corresponding questions using the template-based baseline and MP Triples TransE++.}
\label{table:appendix_examples_2}
\end{table*}

\end{document}